\pdfoutput=1

\documentclass[11pt]{article}

\usepackage{acl}
\usepackage{adjustbox}
\usepackage{times}
\usepackage{latexsym}
\usepackage[utf8]{inputenc} 
\usepackage[T1]{fontenc}    
\usepackage{hyperref}       
\usepackage{url}            
\usepackage{booktabs}       
\usepackage{amsfonts}       
\usepackage{nicefrac}       
\usepackage{microtype}      
\usepackage{lipsum}
\usepackage{fancyhdr}       
\usepackage{graphicx}       
\graphicspath{{media/}}     
\usepackage{makecell}
\usepackage{colortbl}  
\usepackage{xcolor}
\usepackage{array}   
\definecolor{lightblue}{RGB}{212,220,247}
\definecolor{lightred}{RGB}{241,208,207}
\definecolor{darkgreen}{RGB}{50,100,0}
\usepackage{tcolorbox}
\usepackage{pifont}
\usepackage{booktabs}
\usepackage{multirow}
\usepackage{hyperref}
\usepackage{url}
\usepackage{wasysym}
\usepackage{tikz}
\usetikzlibrary{shapes}
\usepackage{pgfplots}

\usepackage{amssymb}
\usepackage{enumerate}
\usepackage{wrapfig}
\usepackage[T1]{fontenc}

\usepackage[utf8]{inputenc}
\usepackage{tabularx}

\usepackage{microtype}

\title{\includegraphics[width=1.4em]{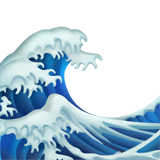}\raisebox{-0.1\height}~
 WaveCoder: Widespread And Versatile Enhancement For Code Large Language Models By Instruction Tuning}

\author{
  Zhaojian Yu$^{1}$\thanks{~~Work was done during an internship at Microsoft.}\;\;\;
  Xin Zhang$^2$\thanks{~~Corresponding author.}\;\;\;
  Ning Shang$^2$ \;\;\;
  Yangyu Huang$^2$\;\;\;
  \textbf{Can Xu}$^2$\;\;\;\\
  \textbf{Yishujie Zhao}$^{1*}$\;\;\;
  \textbf{Wenxiang Hu}$^2$\;\;\;
  \textbf{Qiufeng Yin}$^2$ \;\;\;
  \\
  $^1$Tsinghua University \\
  $^2$Microsoft \\
  \texttt{yzj23@mails.tsinghua.edu.cn}, \texttt{xinzhang3@microsoft.com}\\
  {\url{https://github.com/microsoft/WaveCoder}}
  }

\begin{document}
\maketitle
\begin{abstract}
Recent work demonstrates that, after instruction tuning, Code Large Language Models (Code LLMs) can obtain impressive capabilities to address a wide range of code-related tasks. 
However, current instruction tuning methods for Code LLMs mainly focus on the traditional code generation task, resulting in poor performance in complex multi-task scenarios. 
In this paper, we 
concentrate on multiple code-related tasks and
present \textbf{WaveCoder}, a series of Code LLMs trained with \underline{W}idespread \underline{A}nd \underline{V}ersatile \underline{E}nhanced instruction data.
To enable the models to tackle complex code-related tasks,  we propose a method to stably generate diverse, high-quality instruction data from open source code dataset in multi-task scenarios and obtain \textbf{CodeSeaXDataset}, a dataset  comprising 19,915 instruction instances across 4 code-related tasks, which is aimed at improving the generalization ability of Code LLM.
Our experiments demonstrate that WaveCoder models significantly outperform other open-source models in terms of the generalization ability across different code-related tasks. 
Moreover, WaveCoder-Ultra-6.7B presents the state-of-the-art generalization abilities on a wide range of code-related tasks.


\end{abstract}

\section{Introduction}

\begin{figure*}[t!]
  \centering
  \includegraphics[width=\textwidth]{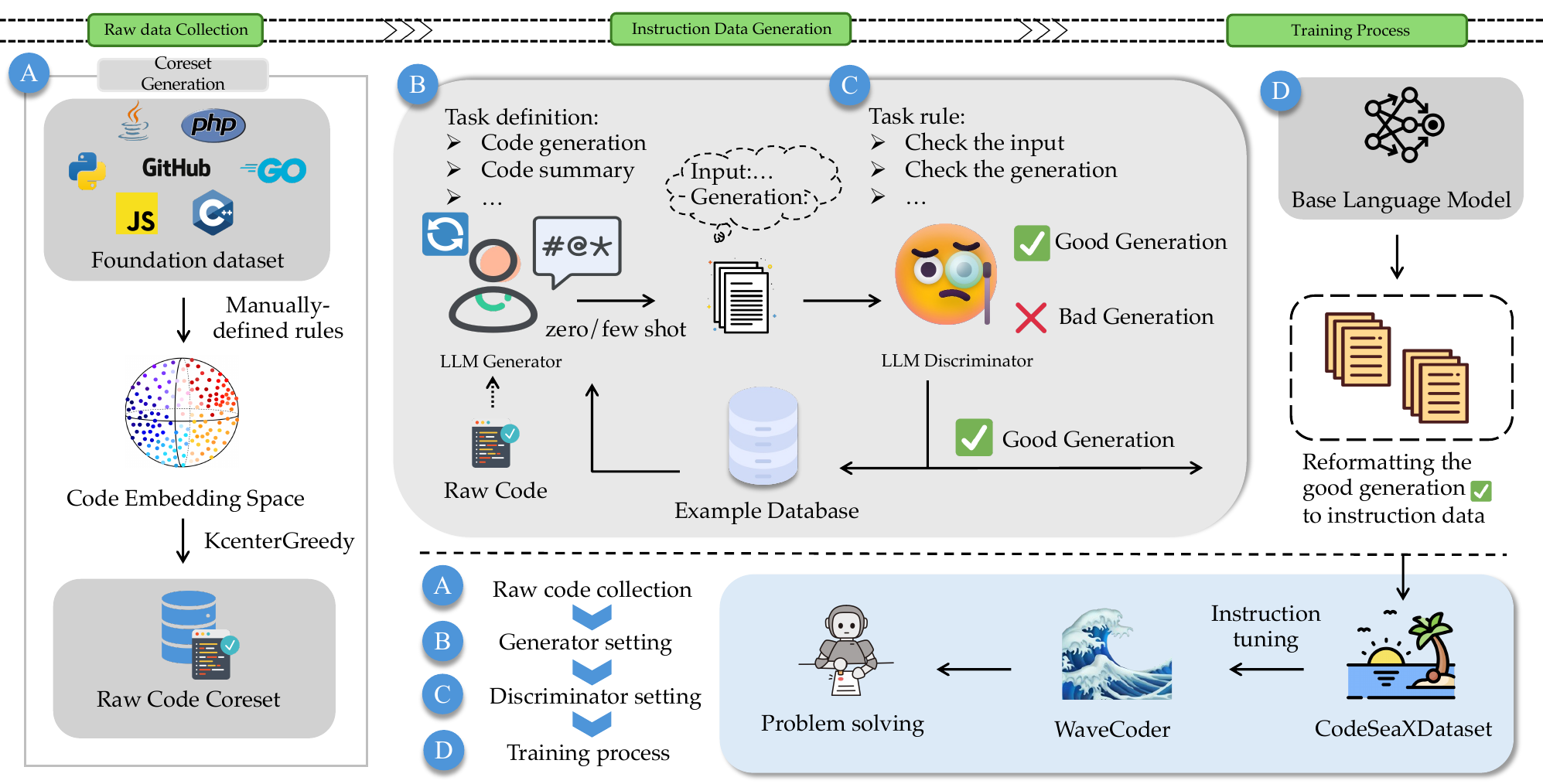}
  \caption{The overview of the widespread and versatile enhancement for Code LLM.
  Part B and C indicates the LLM-based Generator and LLM-based Disciminator where 
  the generator can leverage different examples in example database by in-context learning.
  }
  \label{pipeline}
\end{figure*}



Recently, Large Language Models (LLMs) such as ChatGPT, GPT-4 \cite{openai2023gpt4}, and Gemini \footnote{\url{https://deepmind.google/technologies/gemini}} have attained unprecedented performance levels in a broad array of NLP tasks. These models utilize a self-supervised pre-training process, and subsequent supervised fine-tuning to demonstrate exceptional zero/few-shot capabilities, effectively following human instructions across various tasks.


For code-related tasks,
several previous works, including Codex \cite{chen2021evaluating},
StarCoder \cite{li2023starcoder},  CodeLLaMa \cite{roziere2023code} and DeepseekCoder~\cite{deepseek-coder}, have successfully demonstrated that pre-training on code corpus can significantly improve the model's capability to tackle code-related problems.
After the process of pre-training,
instruction tuning \cite{wei2022finetuned,aribandi2022ext,chung2022scaling} has shown its effectiveness in the aspect of improving the quality of LLM responses.
To specifically enhance the performance of Code LLMs on code-related tasks through instruction tuning, many existing methods for instruction data generation have been designed.
For example, Code Alpaca \cite{codealpaca} utilizes the method of self-instruct~\cite{wang-etal-2023-self-instruct} within the coding domain, leveraging the few-shot capabilities of teacher LLM to generate instruction data.
Similarly, WizardCoder \cite{luo2023wizardcoder} applies the evol-instruct~\cite{xu2023wizardlm} approach based on Code Alpaca, demonstrating a novel and effective method for the generation of instruction data.
These applications underscore the potential of utilizing teacher LLMs to produce instructional content effectively, thereby offering an avenue for the creation of instruction data in the code domain.
However, the quality of the data they generate  heavily relies
on the performance of the teacher LLM and the limited initial seeds, which often produces a large amount of duplicate instruction instances and reduce the effectiveness of instruction tuning \cite{xu2022learning,yan2023understanding,lee2022deduplicating}.
To break away from dependence on teacher LLMs,
Octopack \cite{muennighoff2023octopack} constructs a code instruction dataset leveraging the natural structure of Git commits.
Nonetheless, ensuring the quality of data in git messages presents a considerable challenge, and the comprehensive screening of data through artificial filtering rules is often a complex task.
Additionally, these endeavors are predominantly centered on traditional code generation tasks and lack the capability to produce detailed, task-specific instructions in multi-task scenarios.

In this paper, we primarily focus on multiple code-related tasks, aiming to generate high-quality and diverse instructional data tailored to specific task requirements. Addressing the aforementioned challenges, we refine the instruction data by classifying the instruction instances to four universal code-related tasks in CodeXGLUE 
 \cite{lu2021codexglue}: 1) Code Summarization, 2) Code Generation, 3) Code Translation, 4) Code Repair and propose a  widespread and versatile enhanced instruction generation method  that could make full use of open source code data and stably generate high quality and diverse instruction data  in  multi-task scenarios.
By this generation strategy,
we obtain a dataset of 19,915 instruction instances across four code-related tasks, termed \textbf{CodeSeaXDataset}.

To validate our approach, we
train StarCoder \cite{li2023starcoder}, CodeLLaMa \cite{roziere2023code}, and DeepseekCoder \cite{deepseek-coder} with our initial CodeSeaXDataset dataset and get 
\textbf{WaveCoder}.
Following a thorough assessment on HumanEval \cite{chen2021evaluating}, MBPP \cite{austin2021program}, HumanEvalPack \cite{muennighoff2023octopack} benchmarks,  experimental results show that our \textbf{WaveCoder} exhibits outstanding generalization ability based on widespread and versatile enhanced instruction tuning. 
Moreover, to further explore the improvements brought by data quality, we use GPT-4 \cite{openai2023gpt4} to regenerate response for the instruction in CodeSeaXDataset. Fine-tuned with the enhanced 20K CodeSeaXDataset dataset, we obtain \textbf{WaveCoder-Pro-6.7B} which achieve 72.0\% pass@1 on HumanEval~\cite{chen2021evaluating} and surpass open source Code LLMs but still behind SoTA Code LLM. 
Combining enhanced CodeSeaXDataset with WaveCoder-evol-codealpaca, the decontaminated  Magicoder-evol-codealpaca \footnote{\url{https://huggingface.co/datasets/ise-uiuc/Magicoder-evol-codealpaca-110K}} dataset,
we present \textbf{WaveCoder-Ultra-6.7B}, with SoTA generalization capabilities on multiple code-related tasks. 

\section{CodeSeaXDataset: Four-task Code-related  Instruction Data}


\subsection{Tasks Details}
\begin{table*}[t]
\centering
\caption{The proportion of generated data in generation phase.}
\label{prompts}
\small
\begin{tabular}{llcl}
\toprule

\textbf{Task}               & \textbf{Num}   & \textbf{Per(\%)} & \textbf{Prompt}                                                                                                 \\
\midrule
Code Generation    & 11370 & 57.1           & Implementing functions that perform specific operations given input. \\
Code Summarization & 3165  & 15.8            & Write clear and concise documentation for the given code.  \\
Code Repair        & 3144  & 15.8 
&Identify and fix errors in the given code.   \\
Code Translation   & 2236  & 11.2     &Rewrite the given code from one programming language to another.      \\
\bottomrule
\end{tabular}

\end{table*}
Given the code-related tasks from CodeXGLUE \cite{lu2021codexglue}, we select four of the most universally representative and common tasks from the three generative tasks (code-to-text, text-to-code, and code-to-code) for further exploration including Code Summarization, Code Generation, Code Translation, and Code Repair. Detailed descriptions of these tasks can be found below.

\begin{table}[t]

\centering
\caption{The proportion of  programming language  in raw code.}
\label{pl}
\small
\begin{tabular}{lc}
\toprule
\textbf{Task   }               & \textbf{Percentage(\%)} \\
\midrule
Python                & 29.44          \\
PHP                   & 21.34          \\
Go                    & 19.68          \\
Java                  & 18.53          \\
JavaScript            & 5.56           \\
Others (Ruby,C++,C\#) & 5.45    \\
\bottomrule
\end{tabular}

\end{table}
\noindent\textbf{Code Summarization (code-to-text).} This task aims to create a brief summary of a given code. The raw code is used as input and the teacher model's response is reformulated into an instruction format.





\noindent\textbf{Code Generation (text-to-code, code-to-code).} 
In this task, the model is expected to generate code based on a user's demand description. Therefore, the teacher model is expected to generate instructions and solution code given the raw code as a instruction-solution pair. The generated solution code is then considered as the output.

\noindent\textbf{Code Translation (code-to-code).} 
This task involves converting one programming language into another. The task-specific prompt and raw code are given to the teacher model, then the model generates instructions and the translated code.

\noindent\textbf{Code Repair (code-to-code).} 
The aim of this task is to provide correct code based on potential issues in the given code. The teacher model is expected to generate solutions for the incorrect code, typically with the correct code and some descriptions, which are then taken as the output.

\subsection{Widespread and Versatile Enhanced Instruction Generation}
In past research work \cite{zhou2023lima,gupta2023instruction}, many researchers have discovered that data quality and diversity often play a more important role in instruction tuning process than data amount. The improvement of data quality and diversity are directly related to the performance of the fine-tuned LLM. Therefore, to ensure the data quality and diversity of instruction instance,  we propose a  widespread and versatile enhanced instruction generation method including two the following parts: 
1) a method that can retain the diversity of instruction data
by retainig the diversity of raw code to the utmost extent. 
2) a LLM-based Generator-Discriminator framework to stably generate high-quality instruction data.

\begin{figure*}[t!]
  \centering
  \includegraphics[width=\textwidth]{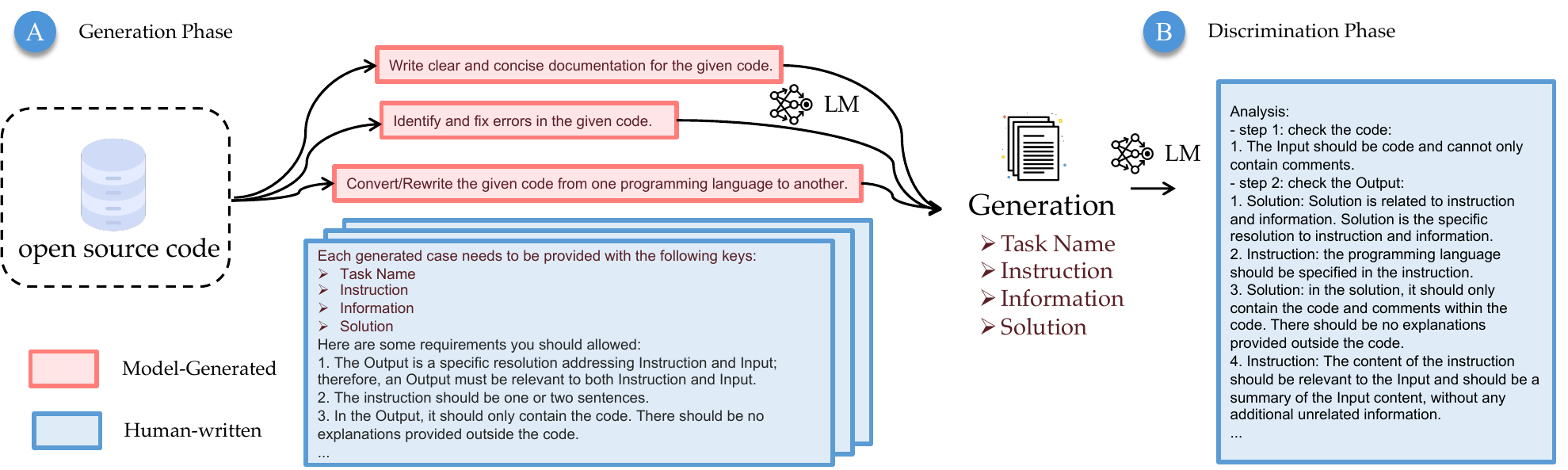}
  \caption{The overview of the our LLM-based Generator-Discriminator framework. In part A, the output of Generator includes 4 keys: Task name, Instruction, Information, Solution. All keys will be analyzed in the Discrimination Phase and the analysis can be reused as examples in next turn.}
  \label{datageneration}
\end{figure*}

\subsubsection{Raw Code Collection}
To ensure the quality and diversity of raw code, we manually define some filtering rules and utilize a cluster method KCenterGreedy \cite{sener2018active,chen2023maybe} to get the raw code collection from the open source code dataset.
In this work, we select CodeSearchNet \footnote{\url{https://huggingface.co/datasets/code_search_net}}, which contains 2 million of <comment, code> pairs from open-source libraries hosted on GitHub, as our foundation dataset and process it with the following steps: 


\noindent\textbf{Manually defined filtering rules.} In order to select high-quality code for instruction-tuning, we make the following rules to filter the foundation dataset: 
i) In this work, we filtered the code to
make sure that the length of required code is neither too long nor too short.
ii) Followed Code Alpaca~\cite{codealpaca}, we have eliminated the raw code containing words from the blacklist, which could potentially reduce the performance of the resulting model.

\noindent\textbf{Coreset selection method.}
To ensure the data diversity when select raw code samples,
we employed KCenterGreedy  \cite{sener2018active} algorithm, which has been proven efficient in obtaining a set of core samples of one distribution, to select representative samples from the open source code dataset based on the code embeddings encoded by the same embedding model (\texttt{roberta-large-v1} \cite{liu2019roberta}).


By incorporating such a method into the open source code dataset, the diversity of the generated data no longer relies solely on capability of the teacher LLM itself or initial seed.
Moreover, due to the application of the KCenterGreedy algorithm, the diversity of languages is also significantly retained, as shown in Table \ref{pl}.





\begin{table*}[t]
\definecolor{deepgreen}{RGB}{0,100,0}
  \centering
  \caption{Results of pass@1 on HumanEval and MBPP benchmark. We use self-reported scores whenever available. 
    The abbreviations "CL", "SC", "DS" refer to the base models CodeLLaMa and StarCoder and DeepseekCoder, respectively. "WaveCoder-Pro-6.7B" and "WaveCoder-Ultra-6.7B" is detailed in the last paragraph of Section 1.
  Due to the difference in decoding strategies from previous evaluation work, we marked the results of greedy decoding in \colorbox{lightblue}{blue} and n = 200 samples in \colorbox{lightred}{red}. -: Not reported in their paper.
 }
    \small
  \label{tab1}
  \renewcommand{\arraystretch}{0.95}
  \begin{tabular}{lr|cr|cc}
    \toprule
    
    \textbf{Model} & \textbf{Params}  & \textbf{Base Model} & \textbf{InsT Data}   & \textbf{HumanEval} & \textbf{MBPP (500)} \\
    \midrule
     \multicolumn{6}{c}{Proprietary Models}  \\
    \midrule

    GPT-4 & - & - & -  &\colorbox{lightblue}{85.4} / \colorbox{lightred}{67.0} & - \\   
    ChatGPT  & - & - & -  & \colorbox{lightblue}{73.2} / \colorbox{lightred}{48.1} & \cellcolor{lightred}{52.2} \\ 


    
    \midrule
    \multicolumn{6}{c}{Open-Source Models}  \\

    \midrule 
    StarCoder & 15B & - & \color{red}\ding{56}  &  \cellcolor{lightred}{33.6} & \cellcolor{lightred}{43.3} \\
    {OctoCoder} & 15B & StarCoder & 13K   &  \cellcolor{lightred}{46.2} & \cellcolor{lightred}{43.5}  \\
    {WizardCoder } & 15B & StarCoder & 78K   &  \cellcolor{lightred}{57.3} & \cellcolor{lightred}{51.8}  \\
    
    WaveCoder-SC-15B  \raisebox{-0.1\height}{\includegraphics[width=1.2em]{wave.png}} & 15B& StarCoder & 20K& \cellcolor{lightred}{\textbf{50.5 (+16.9)}}& \cellcolor{lightred}{\textbf{51.0 (+7.4)}} \\
    \midrule
    {CodeLLaMa}& 7B& - & \color{red}\ding{56}    & \cellcolor{lightblue}{33.5} & \cellcolor{lightblue}{41.4} \\ 
     {CodeLLaMa-instruct}& 7B & CodeLLaMa & 14K  & \cellcolor{lightblue}{34.8}  & \cellcolor{lightblue}{44.4}\\ 
    WaveCoder-CL-7B  \raisebox{-0.1\height}{\includegraphics[width=1.2em]{wave.png}} & 7B& CodeLLaMa & 20K & \cellcolor{lightblue}{\textbf{48.1 (+14.6)} } & \cellcolor{lightblue}{\textbf{47.2 (+5.8)} } \\
    \midrule
    {CodeLLaMa}& 13B& -  & \color{red}\ding{56}   & \cellcolor{lightblue}{36.0} & \cellcolor{lightblue}{47.0} \\ 
     {CodeLLaMa-instruct}& 13B& CodeLLaMa & 14K& \cellcolor{lightblue}{42.5} & \cellcolor{lightblue}{49.4}\\ 

    WaveCoder-CL-13B   \raisebox{-0.1\height}{\includegraphics[width=1.2em]{wave.png}}& 13B& CodeLLaMa & 20K  & \cellcolor{lightblue}{\textbf{55.4 (+19.4)}}& \cellcolor{lightblue}{\textbf{49.6 (+2.6)}} \\
    \midrule
    DeepseekCoder & 6.7B& - & \color{red}\ding{56} & \cellcolor{lightblue}{49.4} & \cellcolor{lightblue}{60.6} \\
    Magicoder-DS & 6.7B& DeepseekCoder & 75K & \cellcolor{lightblue}{66.5} & \cellcolor{lightblue}{60.4} \\     

    WaveCoder-DS-6.7B  \raisebox{-0.1\height}{\includegraphics[width=1.2em]{wave.png}}& 6.7B& DeepseekCoder & 20K & \cellcolor{lightblue}{\textbf{64.0 (+14.6)}} & \cellcolor{lightblue}{\textbf{62.8 (+2.2)}}\\
    
    WaveCoder-Pro-6.7B  \raisebox{-0.1\height}{\includegraphics[width=1.2em]{wave.png}}& 6.7B& DeepseekCoder & 20K & \cellcolor{lightblue}{\textbf{72.0 (+22.6)}} & \cellcolor{lightblue}{\textbf{63.6 (+3.0)}}\\



    \midrule
    \multicolumn{6}{c}{SoTA Open-Source Models}  \\
    \midrule
    DeepseekCoder-instruct$^*$ & 6.7B& DeepseekCoder & - & \cellcolor{lightblue}{73.8} & \cellcolor{lightblue}{62.8} \\  
   Magicoder-S-DS & 6.7B& DeepseekCoder & 185K & \cellcolor{lightblue}{76.8} & \cellcolor{lightblue}{64.6} \\   
    WaveCoder-Ultra-6.7B  \raisebox{-0.1\height}{\includegraphics[width=1.2em]{wave.png}}& 6.7B& 
    DeepseekCoder & 130K & \cellcolor{lightblue}{\textbf{78.6 (+29.2)}} & \cellcolor{lightblue}{\textbf{64.4 (+3.8)}}\\

    \bottomrule
  \end{tabular}
\end{table*}

\subsubsection{LLM-based Generator-Discriminator Framework}

After the process of raw code collection, the data diversity 
from raw code has been retained, where
the next step is to generate instruction data for supervised fine-tuning from the raw code. To further ensure the quality of generated instruction data, shown in Figure \ref{datageneration}, we propose a LLM-based Generator-Discriminator framework where the generator can leverage an extensive amount of unsupervised open source code to generate supervised instruction data and the discriminator can generate analysis for each component in instruction data.

\noindent\textbf{Generation Phase.} In the generation phase, we utilize GPT-4 to generate definitions for each code-related task. 
As shown in Figure \ref{datageneration}, following the model-generated task definition, we manually develop the generation requirements for the each code-related task. Integrating both the task definition and all associated requirements into the generation prompt, we take the raw code as an input and select different examples from the example database to generate instruction data by GPT-3.5. 



\noindent\textbf{Discrimination Phase.} 
During the exploration of the instruction generation process, we noticed that the data quality of the instruction instances cannot be guaranteed through the generation phase alone. In order to enhance the controllability of data generation and further ensure the data quality, we employ GPT-4 as a LLM-based discriminator to continue analyzing and filtering the instruction data.
Subsequently, inspired by Zero-shot-CoT \cite{kojima2022large}, we establish a series of rules, exemplified in Figure \ref{f3} and  disassemble them to some subtopics to ensure the discriminating accuracy where the LLM-based discriminator can analyze the generation step by step. By adopting this method, the discrimination rules can be modified partially to address certain issues.
After the discrimination process, as shown in Figure \ref{pipeline}, each instruction instance is classified as either a good or bad case and the classification information is subsequently random selected in the following generation as examples. For the reusage of these classified instruction instance, different from self-instruct \cite{wang-etal-2023-self-instruct} which solely utilize the initial seed task as good example, we exploit both the good generation and bad generation as few-shot example so that the generator can learn from the mistake in different bad example.
 Therefore, this framework provides a comprehensive approach to generating and evaluating instruction data, ensuring a high-quality  training dataset. 

\begin{table*}[t]
  \centering
  \caption{Results of pass@1 on HumanEvalFix benchmark. We use self-reported scores whenever available.  Due to the difference in decoding strategies from previous evaluation work, we marked the results of greedy decoding in \colorbox{lightblue}{blue} and n = 20 samples in \colorbox{lightred}{red}. }
  \label{humanevalfix}
  \renewcommand{\arraystretch}{0.95}
  \small
  \begin{tabular}{lccccccc}
    \toprule
    \textbf{Model} & \textbf{Python} &  \textbf{JavaScript} &  \textbf{Java} &  \textbf{Go} & \textbf{C++} &  \textbf{Rust} &  \textbf{Avg.}  \\
    \midrule

    GPT-4  & 47.0& 48.2 & 50.0 & 50.6 & 47.6 & 43.3 & 47.8 \\
    \midrule
    StarCoder  & 8.7 & 15.7 & 13.3 & 20.1 & 15.6 & 6.7 & 13.4 \\
    OctoCoder  & 30.4 & 28.4 & 30.6 & 30.2 & 26.1 & 16.5 & 27.0 \\
    WizardCoder  & 31.8 & 29.5 & 30.7 & 30.4 & 18.7 & 13.0 & 25.7 \\
    \rowcolor{lightred}
    WaveCoder-SC-15B  \raisebox{-0.1\height}{\includegraphics[width=1.2em]{wave.png}} & \textbf{39.3} &  \textbf{35.1} &  \textbf{34.8} &  \textbf{36.2} &  \textbf{30.2} &  \textbf{22.5} &  \textbf{33.0}\\
    \midrule
    CodeLLaMa-instruct-7B & 28.0 & 23.2 & 23.2 & 18.3 & 0.1 & 0.1 & 15.5 \\
    CodeLLaMa-CodeAlpaca-7B & 37.8 & 39.0 & 42.0 & 37.8 & 37.2 & 29.2 & 37.1  \\
    \rowcolor{lightblue}
    WaveCoder-CL-7B  \raisebox{-0.1\height}{\includegraphics[width=1.2em]{wave.png}} & \textbf{41.4} &  \textbf{41.4} &  \textbf{42.0} &  \textbf{47.1} &  \textbf{42.7} &  \textbf{34.7} &  \textbf{41.5}\\
    \midrule
    CodeLLaMa-instruct-13B & 29.2 & 19.5 & 32.3 & 24.4 & 12.8 & 0.1 & 19.7 \\
    CodeLLaMa-CodeAlpaca-13B & 42.7 & 43.9 & 50.0 & 45.7 & 39.6 & 37.2 & 43.2\\
    \rowcolor{lightblue}
    WaveCoder-CL-13B  \raisebox{-0.1\height}{\includegraphics[width=1.2em]{wave.png}} & \textbf{48.8} &  \textbf{48.2} &  \textbf{50.6} &  \textbf{51.8} &  \textbf{45.1} &  \textbf{40.2} &  \textbf{47.4}\\
    \midrule
    DeepseekCoder-6.7B & 29.9 & 29.2 & 39.0 & 29.2 & 25.0 & 21.9 & 29.0 \\
    Magicoder-DS & 42.0 & 43.3 & 50.6 & 41.4 & 38.4 & 29.2 & 40.8 \\

    DeepseekCoder-CodeAlpaca-6.7B & 49.4 & 51.8 & 45.1 & 48.8 & 44.5 & 31.7 & 45.2 \\
    \rowcolor{lightblue}
    WaveCoder-DS-6.7B  \raisebox{-0.1\height}{\includegraphics[width=1.2em]{wave.png}} & \textbf{57.9} & \textbf{52.4} & \textbf{57.3} & \textbf{47.5} & \textbf{45.1} & \textbf{36.0} & \textbf{49.4} \\
    
    \rowcolor{lightblue}
    WaveCoder-Pro-6.7B \raisebox{-0.1\height}{\includegraphics[width=1.2em]{wave.png}} & \textbf{59.1} & \textbf{56.7} & \textbf{54.2} & \textbf{45.1} & \textbf{45.7} & \textbf{34.1} & \textbf{49.2} \\
    \midrule
    Deepseek-instruct-6.7B   & 56.1 & 58.5 & 57.3 & 49.4 & 45.1 & 36.6 & 50.5 \\
    Magicoder-S-DS & 56.1 & 55.4 & 58.5 & 51.2 & 45.7 & 35.3 & 50.3 \\
    \rowcolor{lightblue}
    WaveCoder-Ultra-6.7B  \raisebox{-0.1\height}{\includegraphics[width=1.2em]{wave.png}} & \textbf{58.5} & \textbf{57.3} & \textbf{61.0} & \textbf{53.0} & \textbf{50.0} & \textbf{37.2} & \textbf{52.8} \\

    
    \bottomrule
  \end{tabular}
\end{table*}

\section{Experiments}

\subsection{Setup}
Unlike the previous work \cite{luo2023wizardcoder,shen2023pangu,gunasekar2023textbooks} that mainly focus on code generation task, 
we generate about 20K dataset covers 4 common code-related tasks to enhance the geralization abilties of Code LLM. To obtain WaveCoder models, We choose \textit{StarCoder-15B}, \textit{CodeLLaMa (7B and 13B)}, \textit{DeepseekCoder-6.7B} as the base model and fine-tune all the base model for 3 epochs using  NVIDIA A100-80GB GPU.
For  \textit{StarCoder-15B}, \textit{CodeLLaMa-7B} and \textit{CodeLLaMa-13B}, we set the global batch size to 256 using Tensor Parallel and set the initial learning rate at 2e-5.
For \textit{DeepseekCoder-6.7B}, we set the  global batch size to 512  using  the Fully Sharded Data Parallel (FSDP) module from Pytorch and set the initial learning rate at 5e-5.


\noindent\textbf{Benchmarks and Baselines.}
To ensure a thorough assessment of the model’s generalization ability, we score our model on three code benchmarks across different code related tasks: HumanEval \cite{chen2021evaluating}, MBPP \cite{austin2021program} and HumanEvalPack \cite{muennighoff2023octopack}, as illustrated in Appendix \ref{benchmark}. 

\noindent\textbf{Proprietary Models.} We present the self-reported results from an array of SoTA LLMs, including ChatGPT (\texttt{gpt-3.5-turbo}), GPT-4. If not reported, we use the results from Octopack \cite{muennighoff2023octopack} or evaluate by ourselves.

\noindent\textbf{Open Source Models.} To ensure an equitable comparison, we opted to select models that have been trained with the similar amount of instruction instances for our comparative analysis.

\noindent\textbf{SoTA Open Source Models.}
We compared WaveCoder-6.7B with the SoTA open source Code LLM, includes Magicoder-S-DS \cite{wei2023magicoder} and DeepseekCoder-instruct-6.7B \cite{wei2023magicoder} on a wide range of code-related tasks. All the result of SoTA open source models is presented from EvalPlus. \cite{liu2023is} 
If not reported, we evaluate it by ourselves.



\begin{table*}[t]
  \centering
  \caption{Results of pass@1 on HumanEvalExplain benchmark. We use self-reported scores whenever available. Due to the difference in decoding strategies from previous evaluation work, we marked the results of greedy decoding in \colorbox{lightblue}{blue} and n = 20 samples in \colorbox{lightred}{red}. }
  \label{humanevalexp}
  \renewcommand{\arraystretch}{0.95}
  \small
  \begin{tabular}{lccccccc}
    \toprule
    \textbf{Model} & \textbf{Python} &  \textbf{JavaScript} &  \textbf{Java} &  \textbf{Go} & \textbf{C++} &  \textbf{Rust} &  \textbf{Avg.}  \\
    \midrule

    GPT-4  & 64.6& 57.3 & 51.2 & 58.5 & 38.4 & 42.7 & 52.1 \\

    \midrule
    StarCoder  & 0.0 & 0.0 & 0.0 & 0.0 & 0.0 & 0.0 & 0.0 \\
    WizardCoder  & 32.5 & 33.0 & 27.4 & 26.7 & 28.2 & 16.9 & 27.5 \\

    OctoCoder  & 35.1 & 24.5 & 27.3 & 21.1 & 24.1 & 14.8 & 24.5 \\

   \rowcolor{lightred}
    WaveCoder-SC-15B   \raisebox{-0.1\height}{\includegraphics[width=1.2em]{wave.png}} & \textbf{37.1} &  \textbf{33.3} &  \textbf{40.5} &  \textbf{23.3} &  \textbf{31.8} &  \textbf{19.3} &  \textbf{30.8}\\
   \midrule
    CodeLLaMa-instruct-7B & 33.5 & 36.0 & 31.7 & 21.3 & 25.0 & 16.4 & 27.3 \\
    CodeLLaMa-CodeAlpaca-7B  & 34.7 & 24.4 & 37.8 & 23.2 & 28.6 & 19.5 & 28.0 \\
    \rowcolor{lightblue}
    WaveCoder-CL-7B     \raisebox{-0.1\height}{\includegraphics[width=1.2em]{wave.png}} & \textbf{41.4} &  \textbf{31.7} &  \textbf{39.0} &  \textbf{25.0} &  \textbf{34.1} &  \textbf{23.2} &  \textbf{32.4}\\
    \midrule
    
    CodeLLaMa-instruct-13B & 40.2 & 26.8 & 37.2 & 22.5 & 28.0 & 14.6 & 28.2 \\
    CodeLLaMa-CodeAlpaca-13B  & 32.3 & 28.0 & 34.1 & 18.9 & 29.9 & 20.7 & 27.3 \\
    \rowcolor{lightblue}
    WaveCoder-CL-13B   \raisebox{-0.1\height}{\includegraphics[width=1.2em]{wave.png}} & \textbf{45.7} &  \textbf{42.0} &  \textbf{48.2} &  \textbf{32.3} &  \textbf{38.4} &  \textbf{20.7} &  \textbf{37.9}\\
    \midrule
    
    DeepseekCoder-6.7B & 43.9 & 40.2 & 37.8 & 29.2 & 34.1 & 22.5 & 34.6 \\
    Deepseek-CodeAlpaca-6.7B  & 40.8 & 37.2 & 42.1 & 29.9 & 31.7 & 22.5 & 34.0 \\
    
    Magicoder-DS & 55.5 & 36.6 & 49.4 & 36.0 & 39.6 & 27.4 & 40.7 \\  
    \rowcolor{lightblue}
    WaveCoder-DS-6.7B   \raisebox{-0.1\height}{\includegraphics[width=1.2em]{wave.png}} & \textbf{48.2} &  \textbf{47.5} &  \textbf{49.4} &  \textbf{32.3} &  \textbf{48.2} &  \textbf{22.0} &  \textbf{41.3}\\
    \rowcolor{lightblue}
    WaveCoder-Pro-6.7B   \raisebox{-0.1\height}{\includegraphics[width=1.2em]{wave.png}} & \textbf{53.0} &  \textbf{43.3} &  \textbf{54.9} &  \textbf{34.1} &  \textbf{42.7} &  \textbf{20.0} &  \textbf{41.3}\\
    \midrule
    Magicoder-S-DS & 60.3 & 46.3 & 54.3 & 38.4 & 48.1 & 29.2 & 46.1 \\
     Deepseek-instruct-6.7B & 62.2 & 54.3 & 61.0 & 39.6 & 55.5 & 33.5 & 51.0 \\ 
    \rowcolor{lightblue}
    WaveCoder-Ultra-6.7B   \raisebox{-0.1\height}{\includegraphics[width=1.2em]{wave.png}} & \textbf{56.7} &  \textbf{50.0} &  \textbf{54.3} &  \textbf{34.8} &  \textbf{51.2} &  \textbf{36.6} &  \textbf{47.3}\\

    \bottomrule
    
  \end{tabular}
\end{table*}
\subsection{Result}

\noindent\textbf{Evaluation on Code Generation Task.}
HumanEval and MBPP are two representative benchmarks for code generation task, as illustrated in Appendix \ref{benchmark}.
Table \ref{tab1} shows the pass@1 score of different LLMs on both benchmarks. From the results, We have the following observations:

\noindent1) WaveCoder-Pro-6.7B outperforms other open source models with only 6.7B parameters and 20K instruction data. 
Trained with GPT-4 enhanced CodeSeaXDataset dataset, WaveCoder-Pro-6.7B achieve 72.0\% pass@1 and  on HumanEval and  63.6\% on MBPP, surpassing all open source models but still behind proprietary models and the SoTA open source models. 


\noindent 2) Refined and diverse instruction data can significantly improve the efficiency of instruction tuning. 
As delineated in Table \ref{tab1}, WaveCoder demonstrates commendable performance, utilizing a dataset comprising merely about 20K Instruction Tuning Data (InsT Data), which positions it on an equal footing with its contemporaries. Despite a discernible shortfall in the code generation benchmarks relative to WizardCoder~(50.5 vs 57.3) and Magicoder~(64.0 vs 66.5), it is imperative to consider the substantial disparity in the volume of training data. Moreover, it is observed that WaveCoder-pro-6.7B significantly outperforms Magicoder-DS-6.7B~(72.0 vs 66.5), demonstrating the effectiveness of data quality and diversity in instruction tuning.

\noindent\textbf{Evaluation on Other Code-related Task.}
We score WaveCoder with state-of-the-art Code LLMs on HumanEvalPack \cite{muennighoff2023octopack} in Table \ref{humanevalfix} and Table \ref{humanevalexp}, highlighting the the following salient observations: 

\noindent
1) WaveCoder models outperform all open source models on other code-related task.
Building upon Starcoder, our proposed WaveCoder-SC has exhibited exceptional performance, transcending the capabilities of both WizardCoder  and OctoCoder as evidenced by the HumanEvalFix (33.0 vs 25.7 vs 27.0) and HumanEvalExplain (30.8 vs 27.5 vs 24.5) benchmarks, which is also shown in other base models. 
Notably, WaveCoder-DS-6.7B achieves 49.4\% average pass@1 score on HumanEvalFix and 41.3\% on HumanEvalExplain, surpassing all open source models and demonstrating strong generalization capabilities in multi-task scenarios.

\noindent2) The enhancement in data refinement and diversification can markedly bolster the efficacy of instruction tuning in multi-task scenarios.  Such data refinement, coupled with the categorization of instructions into four code-related tasks, has propelled our models to reach an unforeseen generalization capabilities in various code-related tasks.
Remarkably, our WaveCoder-DS-6.7B model outperforms GPT-4 (49.4 vs 47.8) on HumanEvalFix, thereby underscoring the potential of smaller models to achieve near-parity with parameter-heavy models when optimized efficiently.

\noindent\textbf{WaveCoder-Ultra-6.7B.}
\label{app:wavecoder-ultra}
Inspired by Magicoder-S-DS-6.7B~\cite{wei2023magicoder}, we combine CodeSeaXDataset with WaveCoder-evol-codealpaca to a 130K dataset. Fine-tuned with this combination of two datasets, we obtain WaveCoder-Ultra-6.7B. As illustrated in Table \ref{tab1}, \ref{humanevalfix}, \ref{humanevalexp}, WaveCoder-Ultra-6.7B 
has the state-of-the-art generalization abilities on a wide range of code-related tasks, which highlights the significance of our CodeSeaXDataset dataset again and demonstrates the potential of larger datasets. 
\begin{table*}[t]
  \centering
  \caption{Ablation study on different code-related tasks: CG (\textbf{C}ode \textbf{G}eneration), CS (\textbf{C}ode \textbf{S}ummarization), CT (\textbf{C}ode \textbf{T}ranslation), CR (\textbf{C}ode \textbf{R}epair). WaveCoder-DS-6.7B utilizes all 4 code-related tasks.}
  \label{ablation}
  \small
  \renewcommand{\arraystretch}{0.95}
  \begin{tabular}{l|cccc|ccc}
    \toprule
    \textbf{Model} & \textbf{CG} & \textbf{CS} & \textbf{CT}& \textbf{CR} &  \textbf{HumanEval}   &\makecell[c]{\textbf{HumanEval}\\ \textbf{Fix (Avg.)}} &\makecell[c]{\textbf{HumanEval} \\ \textbf{Explain (Avg.)}}   \\
    \midrule
    DeepseekCoder-Base-6.7B & \color{red}\ding{56} &\color{red}\ding{56} & \color{red}\ding{56}&\color{red}\ding{56}  & 49.4 &   29.0 & 34.6  \\
    \rowcolor{lightblue}
    WaveCoder-DS-6.7B \raisebox{-0.1\height}{\includegraphics[width=1.2em]{wave.png}}   & \color{darkgreen}\ding{52} &\color{darkgreen}\ding{52} & \color{darkgreen}\ding{52}&\color{darkgreen}\ding{52}  & \textbf{64.0 (+14.6)} &   \textbf{49.4 (+20.4)} & \textbf{41.3 (+7.3)}  \\
    \midrule
     -Without Repair&\color{darkgreen}\ding{52} &\color{darkgreen}\ding{52} &\color{darkgreen}\ding{52} & \color{red}\ding{56} & 60.9 (-3.1) &   15.7 (-33.7) & 41.2 (-0.1)  \\
     -Without Generation & \color{red}\ding{56} &\color{darkgreen}\ding{52} & \color{darkgreen}\ding{52}&\color{darkgreen}\ding{52}  & 53.6 (-10.4) &   47.4 (-2.0) & 40.5 (-0.8) \\
      -Without Translation&\color{darkgreen}\ding{52} &\color{darkgreen}\ding{52} & \color{red}\ding{56}&\color{darkgreen}\ding{52}  & 60.9  (-3.1)&   49.3 (-0.1) & 41.6 (+0.3)  \\
     -Without Summarization&\color{darkgreen}\ding{52} &\color{red}\ding{56} &\color{darkgreen}\ding{52}&\color{darkgreen}\ding{52}  & 61.5 (-2.5) &   45.6 (-3.8) & 28.4 (-12.9)  \\

    \bottomrule
    
  \end{tabular}
\end{table*}
\section{Ablation and Analysis}
\subsection{Ablation of Code-related Tasks}
To explore the relationship between different tasks, we conduct an ablation study about the task type of instruction data. Using DeepseekCoder-Base-6.7B as our base model and initial 20K CodeSeaXDataset data as our base dataset, we have the following observations from Table \ref{ablation}:

\noindent1) Refined instruction data can significantly improve the generalization ability of pre-trained models without a tradeoff. As shown in Table \ref{ablation}, incorporating all 4 code-related tasks into training data, WaveCoder-DS-6.7B achieves the best performance on benchmark of all tasks. For example, the participation of the Code Repair task yields a considerable average improvement of 33.7\% absolute for HumanEvalFix without any significant decline in other tasks, and even improved by 3.1\% absolute for HumanEval benchmark.

\noindent2) Different tasks can promote each other so that the model can show a generalization ability. From Table \ref{ablation}, we can observe that 
any combination of three tasks resulted in a lower score than all tasks. For example, the addition of the code summarization task offers a modest yet significant average improvement on all benchmarks. Moreover, the absence of any task will cause the score of HumanEval to drop, which also reflects the mutual promotion between different tasks.

\subsection{Discussion about Data Leakage}
\begin{figure}[h]
  \centering
  \includegraphics[width=0.4999\textwidth]{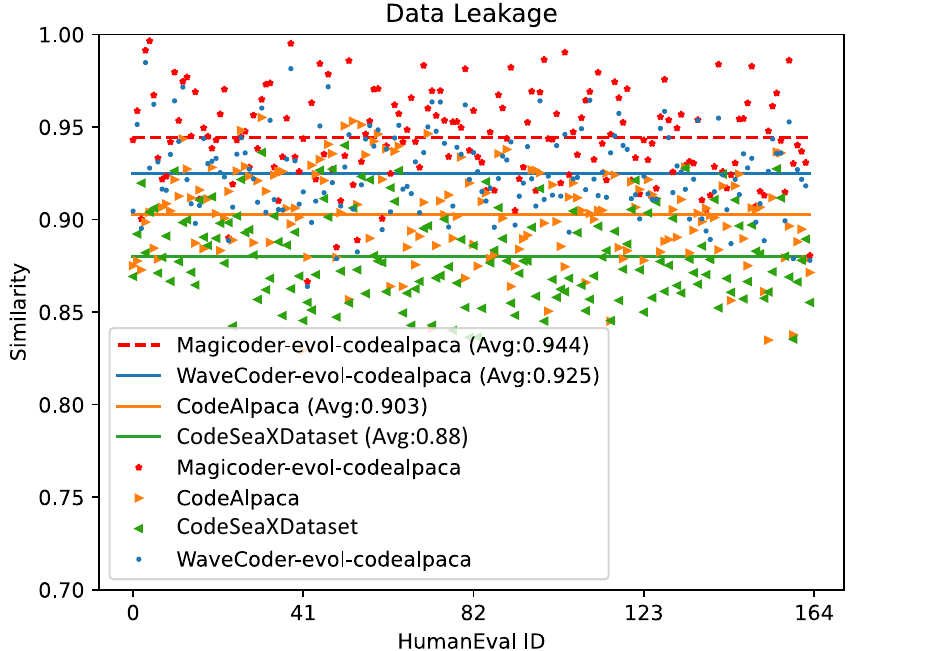}
  \includegraphics[width=0.4999\textwidth]{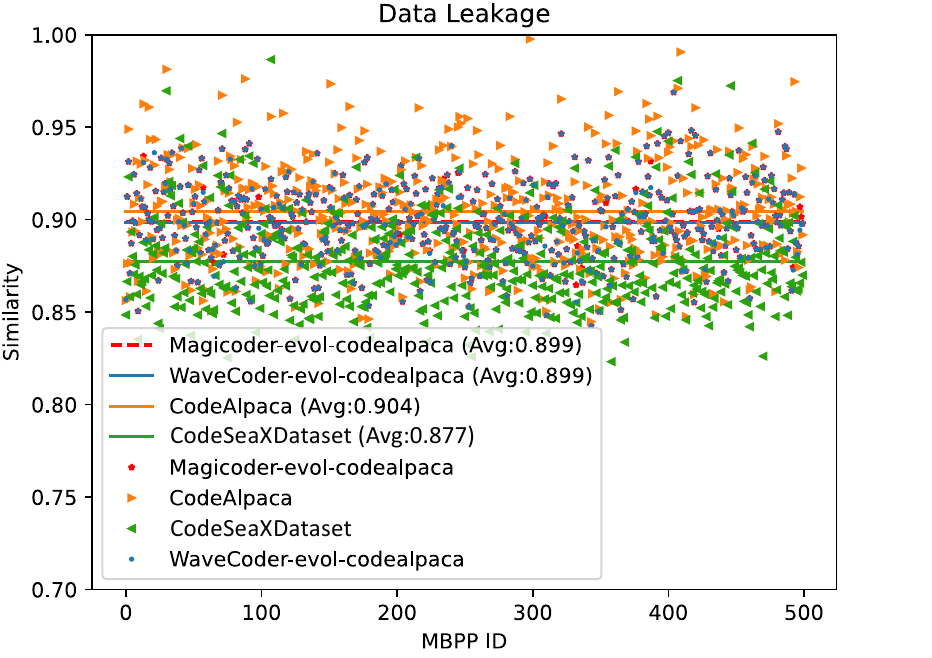}
  \caption{Discussion about data leakage in different training dataset.  
  WaveCoder-evol-codealpaca indicates the decontaminated Magicoder-evol-codealpaca  dataset under our strategy.
  }
  \label{leakage}
\end{figure}

In this section, we explore the potential leakage through three instruction datasets about code~(i.e. Code Alpaca, CodeSeaXDataset, Magicoder-evol-codealpaca). To ensure an accurate analysis, we employ SoTA embedding model \texttt{GTE-Large}~\cite{li2023towards} to encode the canonical code in test benchmarks and all code in training set. Subsequently, we find the nearest neighbour in train set for each questions in test benchmark.
As illustrated in Figure \ref{leakage}, CodeSeaXDataset has the lower average cosine similarity than other datasets. Figure \ref{leakage_exp2} in Appendix presents two examples about the data leakage in these training set.
Moreover, we analyze all benchmarks and notice a serious data leakage issue between HumanEval and Magicoder-evol-codealpaca dataset.
Therefore, we decontaminate Magicoder-evol-codealpaca  
for each evaluation problem in HumanEval and obtain WaveCoder-evol-codealpaca. As illustrated in Figure \ref{leakage}, WaveCoder-evol-codealpaca has lower similarity than Magicoder-evol-codealpaca.

\section{Related Work}
\noindent\textbf{Instruction Tuning.}
Recent studies, such as FLAN \cite{wei2022finetuned}, ExT5 \cite{aribandi2022ext}, and FLANT5 \cite{chung2022scaling}, have underscored the efficacy of integrating diverse tasks within training process to bolster the adaptability of pre-trained models for downstream tasks. Specifically, Flan-PaLM 540B's \cite{chung2022scaling} instruction-tuning over 1.8K tasks has demonstrated that a widespread and versatile enhanced instruction dataset markedly enhances language model performance. InstructGPT  \cite{ouyang2022training}, with its incorporation of premium instruction data crafted by human annotators, has shown significant promise in aligning model outputs with user intents, prompting further investigation into instruction-tuning mechanisms. Additionally, Stanford Alpaca \cite{alpaca} has innovatively employed GPT-generated instruction data via self-instruct \cite{wang-etal-2023-self-instruct} for instruction tuning process. WizardLM \cite{xu2023wizardlm} has built upon these advancements by applying the evol-instruct methodology, collectively illuminating the transformative impact of instruction tuning on the overall capabilities of LLM.

\noindent\textbf{Code Large Language Models.}
Recent advancements in code generation have been propelled by Code LLMs such as CodeGen 
 \cite{nijkamp2022codegen},  CodeT5 \cite{wang2021codet5},  StarCoder ~\cite{li2023starcoder},  CodeLLaMa \cite{roziere2023code} and Deepseek-Coder 
 \cite{deepseek-coder}, which benefit from extensive pre-training on expansive code corpora. Efforts to further enhance efficiency and problem-solving capabilities have led to the development of instruction-tuned models like InstructCodeT5+ \cite{wang2023codet5plus}, WizardCoder  
 \cite{luo2023wizardcoder}, Pangu-coder2 
 \cite{shen2023pangu}, 
However, all the instruction data they used is from Code Alpaca, which is not refined enough in the context of multi-task environment, which drives us to propose new methods for instruction data generation.
Concurrently, with the release of our contemporaneous work Magicoder \cite{wei2023magicoder}, we offer a concise  analysis in Section 3.

\section{Conclusion}
This paper presents WaveCoder, a Code LLM fine-tuned with widespread and versatile enhanced instruction data.
By enabling language models to effectively tackle complex code-related tasks, our approach demonstrates the potential of integrating multiple code-related tasks into instruction tuning for Code LLM and generating high-quality and diverse instruction data for specific task requirements in multi-task scenarios. WaveCoder achieves state-of-the-art generalization performance on different code-related tasks surpassing existing open source Code LLMs. Furthermore, our analysis of the relationship of different tasks provides valuable insights for future research, paving the way for more extensive code-related tasks and larger dataset.

\section*{Limitations}
We present WaveCoder and propose a data generation method which can stably generate high-quality and diversity instruction data from open source dataset in multi-task scenario. One limitation of our work is that the training dataset we used only includes 19,915 instructions, which produces limited enhancements to the model. As illustrated Section 3, we expand the training dataset to a larger amount and the resulted model still have significant improvement. Therefore, future work should focus on more code-related task types and larger dataset.
\section*{Ethics Statement}
We constructed our CodeSeaXDataset dataset from open source code. For each code snippet we used, we are committed to adhering to the terms of its license, which includes proper attribution ensuring that any modifications or derivative works are also shared under the compatible terms. 
Moreover, we notice a serious data leakage issue in the Magicoder-evol-codealpaca dataset. 
To ensure a fair comparison, we remove three nearest neighbours of each question in  test benchmark from train set.
However, if all similar samples are accidentally removed, the integrity of the data will be damaged, which is harmful to model training.
Therefore, this phenomenon should be attributed to the fact that the problems in the current test benchmarks are some of the most basic algorithm logic. To this end, we call for more comprehensive and complex test benchmarks for Code LLMs which will not easily cause data leakage problem.

\bibliography{anthology,custom}
\bibliographystyle{acl_natbib}
\newpage
\onecolumn
\appendix
\section{Prompt}

\label{prompt}
Followed Alpaca \cite{alpaca}, we set the fine-tuning  prompt  as follows:
\definecolor{mycolor}{RGB}{208,223,230}
\begin{tcolorbox}[colback=mycolor]
    \textbf{Prompt with Input:} 
    
        Below is an instruction that describes a task, paired with an input that provides further context. Write a response that appropriately completes the request.
        
        \#\#\# Instruction:\{instruction\}
        
        \#\#\# Input:\{input\}
        
        \#\#\# Response:
    \\[0.3cm]
     \textbf{Prompt without Input:}  
    
        Below is an instruction that describes a task, paired with an input that provides further context. Write a response that appropriately completes the request.
        
        \#\#\# Instruction:\{instruction\}
        
        \#\#\# Response:

\end{tcolorbox}

\section{Examples of the LLM-based Generator-Discriminator framework}
\label{example}
\begin{figure*}[h!]
  \centering
  \includegraphics[width=\textwidth]{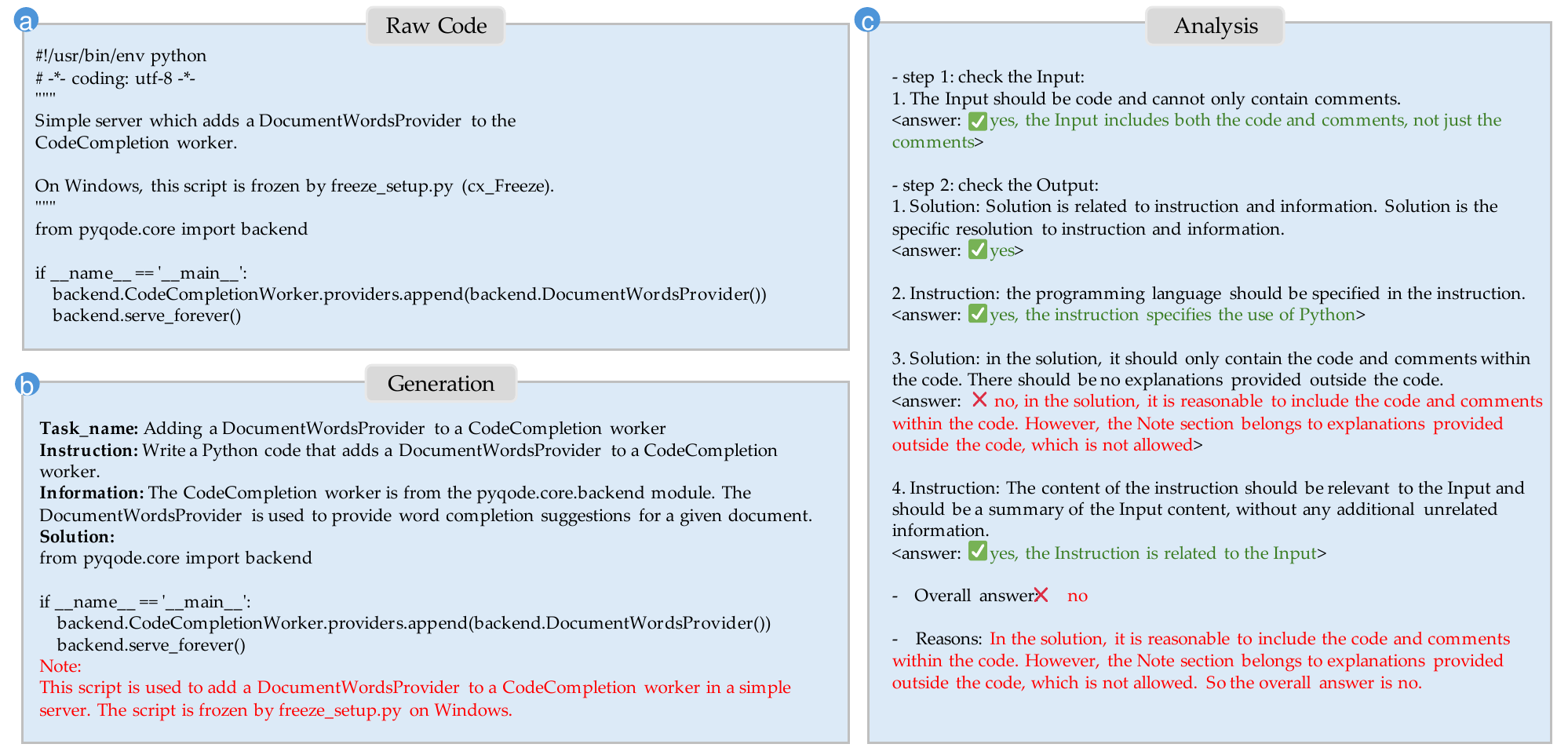}
  \caption{An example of LLM-based Generator-Discriminator framework on code generation task. The generator produce the instruction data based on the input (a). Subsequently, the discriminator accept the output and generated analysis for it. The output (b) includes four keys, we take the information as input and solution as output in our instruction tuning. The analysis (c) consists of the detailed reason for each rule and an overall answer to check if the 
   sample meet all the requirements.
   }
  \label{f3}
\end{figure*}

\begin{table}[h]
\centering
\begin{tabular}{p{0.85\textwidth}}
\toprule
\textbf{LLM-based Generator-Discriminator framework} \\
\midrule
\textbf{Input:}\\
The Python code calculates the area of a circle given its radius.\\
\textbf{Output:}\\
t\textbf{ask\_name:} Calculate Circle Area\\
\textbf{instruction:} Write a Python function that calculates the area of a circle given its radius.\\
\textbf{information:} The formula to calculate the area of a circle is A = pi * $r^2$, where A is the area and r is the radius.\\
\textbf{solution:}
\\
import math

def area\_of\_circle(radius):
    return math.pi * radius ** 2
\\
\textbf{Analysis:}
\textbf{- Step 1: Check the Instruction:} The programming language should be specified in the instruction. <answer: yes, The instruction mentions "Write a Python function," which clearly specifies the programming language as Python.> \\
\textbf{- Step 2: Check the Solution: }\\
  1. Solution is related to instruction and information. Solution is the specific resolution to instruction and information. <answer: yes, The solution provided calculates the area of a circle given its radius, which is directly related to the instruction and the information provided.> \\
  2. In the solution, it should only contain the code and comments within the code. There should be no explanations provided outside the code. <answer: yes, The solution only contains the code without any explanations outside the code. Comments within the code are not present, but the code is straightforward and easy to understand.> \\
  3. The code that contains algorithmic logic in solution should not often be buried inside complex or poorly documented functions, making them difficult to follow or learn from. <answer: yes, The code that contains algorithmic logic in the solution should not often be buried inside complex or poorly documented functions, making them difficult to follow or learn from.> \\
  4. The code should import the required necessary libraries. <answer: yes, The code imports the necessary math library, which is required for calculating the area of a circle using the mathematical constant pi.> \\
\textbf{- Overall answer: }yes \\
\textbf{- Reasons:} All the requirements are met as per the given rules. The programming language is specified, and the solution is related to the instruction and information. The code is simple, easy to follow, and imports the necessary library. The solution provided is a good example of a well-structured and concise code that meets the task's specifications. \\
\bottomrule
\end{tabular}
\caption{An example of the LLM-based Generator-Discriminator framework}
\end{table}

\section{Comparison with CodeAlpaca}
CodeAlpaca dataset contains 20K multi-task instruction-following data generated by the techniques in the self-instruct \cite{alpaca}. To ensure a fair and multidimensional comparison, we randomly sampled 1K and 5K from both datasets (CodeAlpaca and CodeSeaXDataset), set the same set of training hyper-parameters set (epoch = 3, learning rate = 1e-4, LoRA rank = 8) and used the same training prompts. To prevent overfitting, we use Low-Rank 
 Adaption (LoRA) \cite{hu2022lora} for fine-tuning if the size of instruction-follow training dataset is less than 5K and perform full fine-tuning on whole 20K dataset.

\begin{figure*}[h!]
  \centering
  \includegraphics[width=\textwidth]{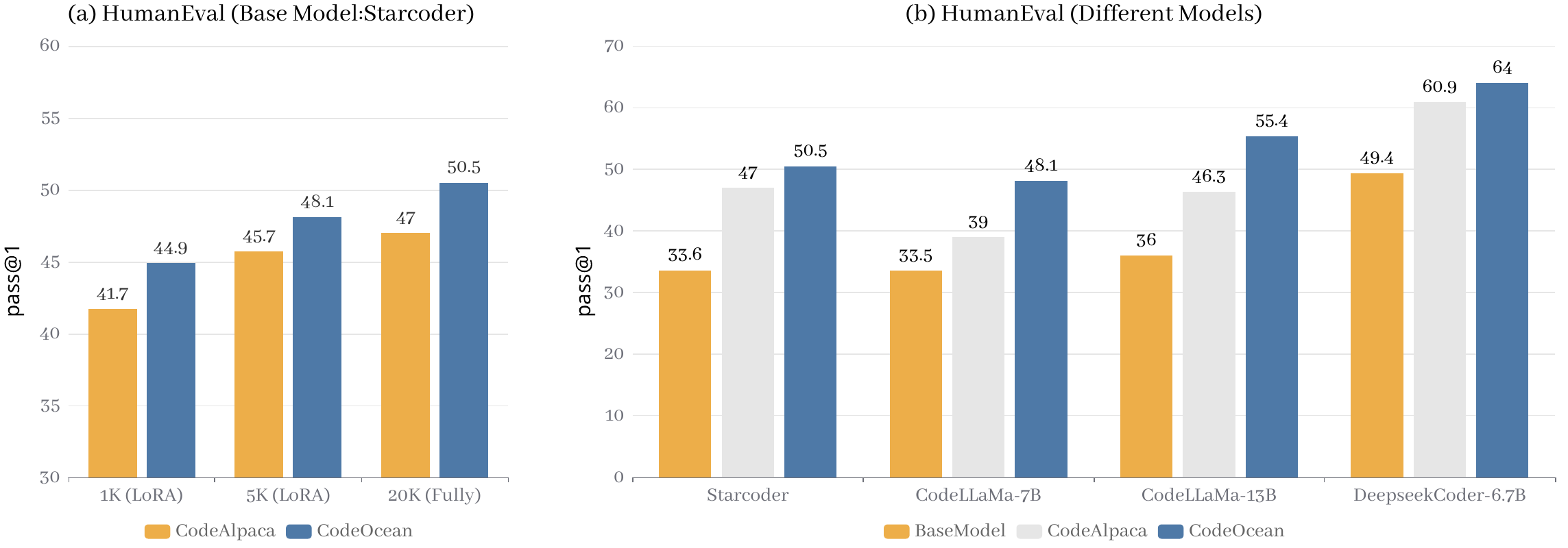}
  \caption{Comparision with CodeAlpaca with different dataset size(a) and different base models(b). CodeSeaXDataset outpeforms CodeAlpaca on HumanEval multidimensionally, more detailed analysis is shown in Section 3.3.}
  \label{biogram}
\end{figure*}

 1) After being fine-tuned with 1K, 5K and 20K of instructional data respectively, the performance of base model improves significantly on HumanEval shown in Figure \ref{biogram}. Taking Starcoder as the base model, CodeSeaXDataset surpasses the CodeAlpaca (44.9\% vs 41.7\%, 45.7\% vs 48.1\% and 47.0\% vs 50.5\%) shown in Figure \ref{biogram} (a), which emphasizes the effectiveness of our method on refining instruction data.
 As shown in Figure \ref{biogram} (b), The results of different base models on CodeSeaXDataset surpasses the results on CodeAlpaca, which emphasizes the effectiveness of CodeSeaXDataset dataset in enhancing the instruction-following ability of the base model.

2) According to Table \ref{humanevalfix} and Table \ref{humanevalexp}, All WaveCoder models significantly outperform the model fine-tuned with CodeAlpaca. Remarkably, The pass@1 score of WaveCoder-CL-13B outperforms CodeLLaMa-CodeAlpaca-13B achieving 10.6\% absolute improvements on HumanEvalExplain. This emphasizes the effectiveness of defining and classifying code-related tasks on enhancing the generalization ability of Code LLMs.
    

\section{Evaluation Benchmark}
\label{benchmark}

\noindent\textbf{HumanEval} \footnote{\url{https://huggingface.co/datasets/openai_humaneval}}, which consists of 164 manually-written Python
programming problems and an average of 9.6 test cases allocated to each problem is now the most 
extensively adopted benchmark for Code LLMs.

\noindent\textbf{MBPP} \footnote{\url{https://huggingface.co/datasets/mbpp}} consists of around 1,000 crowd-sourced Python programming problems, designed to be solvable by entry level programmers, covering programming fundamentals, standard library functionality, and so on. 
In this paper, we choose the 500 problems test dataset to evaluate both few-shot inference of fine-tuned models. For whose MBPP (500) result is not reported or not used, we reproduced for them using bigcode-evaluation-harness \footnote{\url{https://github.com/bigcode-project/bigcode-evaluation-harness}}.

\noindent\textbf{HumanEvalPack }\footnote{\url{https://huggingface.co/datasets/bigcode/humanevalpack}} is an extension of OpenAI's HumanEval to cover 6 total languages across 3 tasks. In this paper, we select the HumanEvalFix to evaluate the code to code ability especially on code repair task and HumanEvalExplain benchmarks to evaluate the code to text ability especially on code summarization task.

\begin{figure*}[t]
  \centering
  \includegraphics[width=\textwidth]{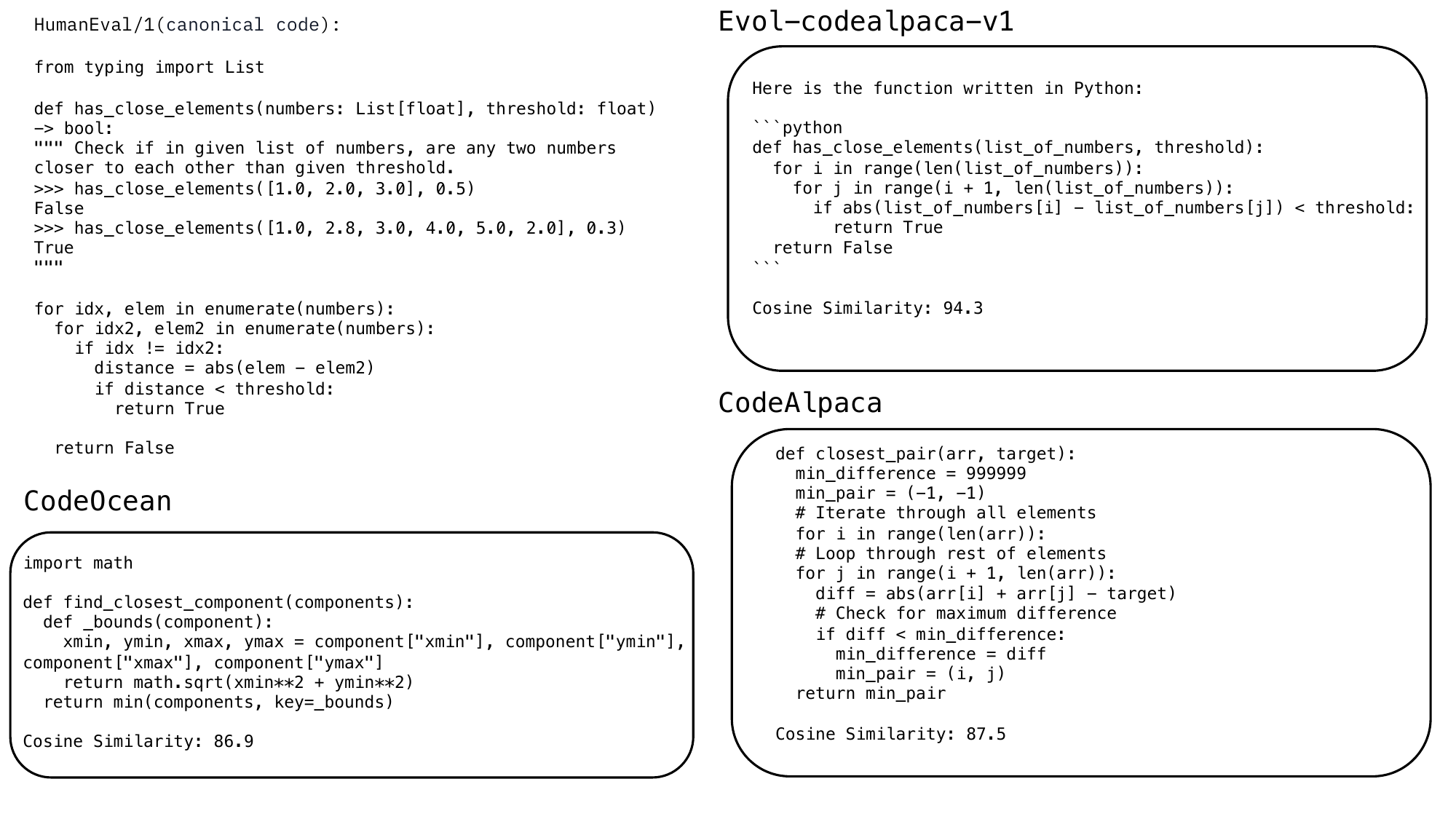}

  \includegraphics[width=\textwidth]{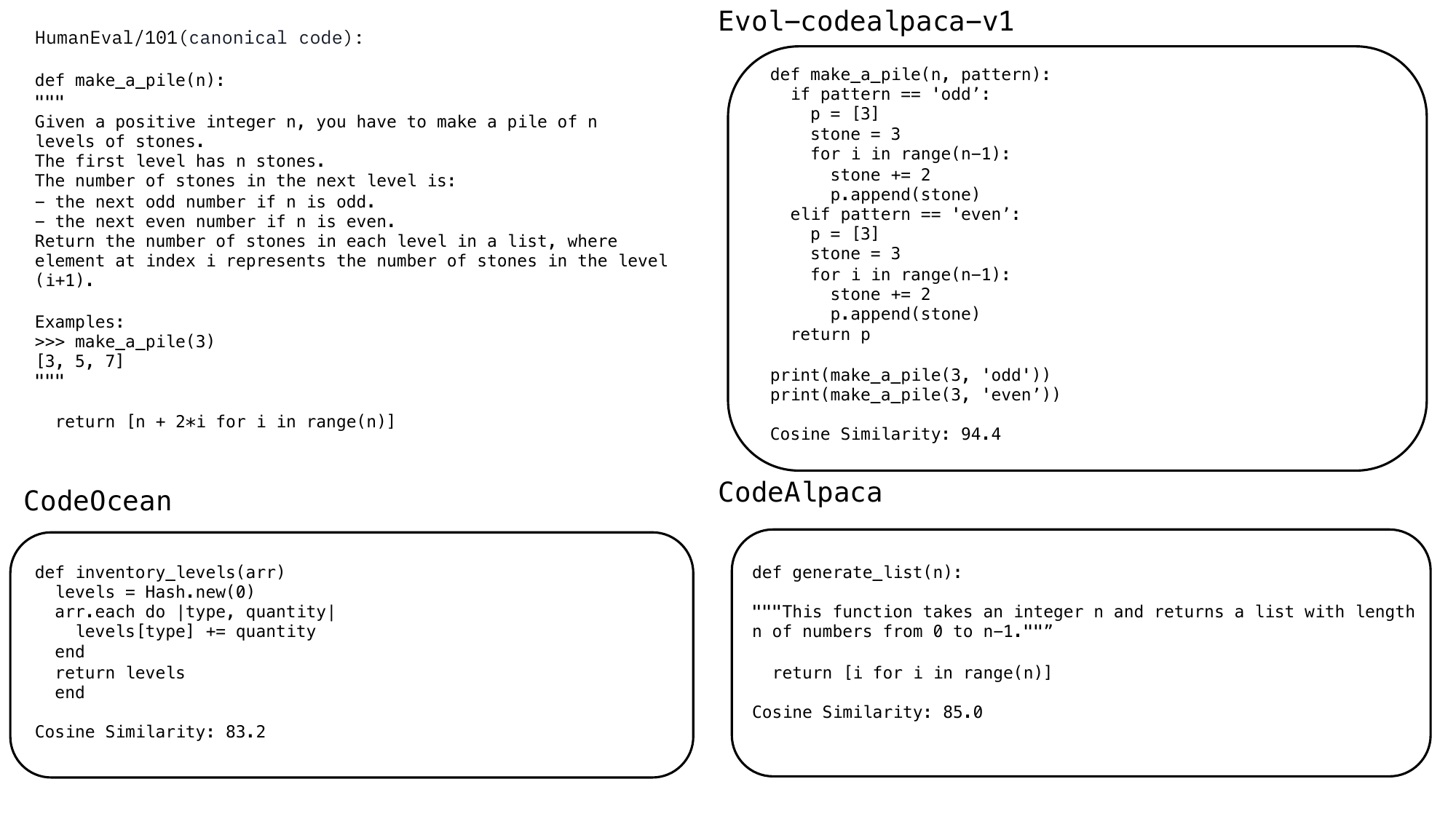}
  \caption{Examples about data leakage.}
  \label{leakage_exp2}
\end{figure*}

\end{document}